\documentclass[
]{ceurart}

\usepackage{listings}

\usepackage{subfigure} 

\usepackage{tikz}
\usepackage{xcolor} 
\usetikzlibrary{shadows.blur, decorations.markings} 
\definecolor{NRM30}{RGB}{31,119,180}
\definecolor{NRM5}{RGB}{255,127,14}
\definecolor{RNN}{RGB}{44,160,44}

\begin{document}

\copyrightyear{2025}
\copyrightclause{Copyright for this paper by its authors.
  Use permitted under Creative Commons License Attribution 4.0
  International (CC BY 4.0).}

\conference{OVERLAY 2025 @ ECAI 2025 October 26th, Bologna (Italy)}

\title{Fully Learnable Neural Reward Machines}
\author[1]{Hazem Dewidar}[%
orcid=0009-0003-2377-2871,
email=hazem.dewidar@uniroma1.it,
url=https://gladia.di.uniroma1.it/authors/dewidar/ ]
\cormark[1]
\fnmark[1]
\address[1]{La Sapienza University of Rome}

\fnmark[1]

\author[1]{Elena Umili}[%
orcid=0000-0002-5639-6038,
email=umili@diag.uniroma1.it,
url=https://sites.google.com/view/elenaumili/home,
]
\fnmark[1]

\cortext[1]{Corresponding author.}
\fntext[1]{These authors contributed equally.}

\begin{abstract}
Non-Markovian Reinforcement Learning (RL) tasks present significant challenges, as agents must reason over entire trajectories of state-action pairs to make optimal decisions. A common strategy to address this is through symbolic formalisms, such as Linear Temporal Logic (LTL) or automata, which provide a structured way to express temporally extended objectives. However, these approaches often rely on restrictive assumptions—such as the availability of a predefined Symbol Grounding (SG) function mapping raw observations to high-level symbolic representations, or prior knowledge of the temporal task. In this work, we propose a fully learnable version of Neural Reward Machines (NRM), which can learn both the SG function and the automaton end-to-end, removing any reliance on prior knowledge. Our approach is therefore as easily applicable as classic deep RL (DRL) approaches, while being far more explainable, because of the finite and compact nature of automata. Furthermore we show that by integrating Fully Learnable Reward Machines (FLNRM) with DRL outperforms previous approaches based on Recurrent Neural Networks (RNNs).
\end{abstract}

\begin{keywords}
Automata Learning \sep
  Neurosymbolic learning \sep
  Deep Reinforcement Learning
\end{keywords}

\maketitle

\section{Introduction}
\noindent
Reinforcement Learning (RL) has achieved remarkable success, yet its foundational assumption of the Markov property—that the present state contains all information needed for optimal decision-making—is often violated in real-world scenarios. Many tasks are inherently non-Markovian, requiring an agent to remember and reason over a history of events to succeed.
The RL community has largely pursued two paths to address non-Markovian tasks. The most common is to equip agents with an RNN ~\cite{ha2018worldmodels, kapturowski2019recurrent}, which learns a compressed representation of history from experience. While powerful, this approach creates a "black box," making the agent's internal reasoning opaque and difficult to verify or trust. An alternative path uses structured, symbolic formalisms like automata or temporal logic~\cite{icarte2022reward, degiacomo2013linear} to explicitly model the task's temporal structure. These models offer outstanding interpretability and support formal verification. However, their practical application is severely limited by a critical bottleneck: they typically presume that both the task automaton and a symbol grounding (SG) function—which maps raw observations like pixels to abstract symbols like `key` or `door`—are provided beforehand~\cite{umili2024neuralrewardmachines}. This reliance on prior, hand-engineered knowledge prevents their use in novel environments where such information is unavailable.
This paper bridges the gap between these two paradigms.
We build on Neural Reward Machines (NRMs) \cite{umili2024neuralrewardmachines}, which use a NeSy architecture to encode automaton knowledge and learn the SG function in a semi-supervised fashion. Here, we show how NRMs can be made fully learnable, thus removing the assumption of a known automaton. We introduce Fully Learnable Neural Reward Machines (FLNRM), the first framework to learn both the task automaton and the symbol grounding function simultaneously and end-to-end, directly from raw sensory inputs and scalar rewards.
 Our core contributions can be summarized as follow: (i) we propose a novel extension of NRM: FLNRM, that eliminates the need for prior knowledge of the task's temporal logic or its symbolic representation, a major limitation of prior automata-based RL methods (ii) we show experimentally that the learned structure provides a powerful inductive bias, enabling FLNRM to outperform standard RNN-based baselines, especially in tasks with complex logical constraints.
Our method therefore retains the general applicability of standard Deep RL approaches, while improving performance and interpretability, taking the best from both automata-based and deep learning-based RL.
\section{Related Works}
Temporal logic formalisms are widely used in Reinforcement Learning (RL) to specify non-Markovian tasks \cite{littman2017}, allowing agents to reason about temporally extended goals and constraints.
Much of the existing literature assumes that: (1) the temporal specification is given, and (2) the boolean propositions used in the specification are observable in the environment—either perfectly \cite{reward-machine-sheila, restr_bolts, reward_machine_learning_1, reward_machine_learning_2, reward_machine_learning_3, subgoalAutomaton} or with some noise \cite{noisy_symbols_2020, noisy_symbols_2022, noisy_symbols_2022_shila}.
Many prior approaches relax only assumption (1), by integrating automata learning within RL agents \cite{reward_machine_learning_1, reward_machine_learning_2, subgoalAutomaton}; or only assumption (2), using neurosymbolic (NeSy) frameworks \cite{umili2024neuralrewardmachines} or multi-task RL techniques \cite{ltl_no_grounding}; yet they still rely on one of the two.

Notably, recent work \cite{hyde2024detectinghiddentriggersmapping} learns both automata and latent event triggers from data without requiring predefined labeling functions or prior temporal knowledge. However, its use of Inductive Logic Programming (ILP) limits applicability only to discrete and finite symbolic domains, excluding environments providing raw observations, such as images or sensor data.
In our approach, we learn the automaton describing the RL task structure directly from raw experience, without any prior knowledge or assumptions about the type of observations—which may be high-dimensional and continuous. 

\section{Background}
\paragraph{Notation}
In this work, we consider \textit{sequential} data of various types, including both symbolic and subsymbolic representations. Symbolic sequences are also called \textit{traces}. Each element in a trace is a symbol $\sigma$ drawn from a finite alphabet $\Sigma$.
We denote sequences using bold notation. For example, $\boldsymbol{\sigma} = (\sigma^{(1)}, \sigma^{(2)}, \ldots, \sigma^{(T)})$ represents a trace of length $T$. Each symbolic variable in the sequence can be grounded either categorically or probabilistically.
In the case of categorical grounding, each element of the trace is assigned a symbol from $\Sigma$, denoted simply as $\sigma^{(i)}$. In the case of probabilistic grounding, each symbolic variable is associated with a probability distribution over $\Sigma$, represented as a vector $\tilde{\sigma}^{(i)} \in \Delta(\Sigma)$, where $\Delta(\Sigma)$ denotes the probability simplex defined as
\[
\Delta(\Sigma) = \left\{ \tilde{\sigma} \in \mathbb{R}^{|\Sigma|} \,\middle|\, \tilde{\sigma}_j \geq 0,\ \sum_{j=1}^{|\Sigma|} \tilde{\sigma}_j = 1 \right\}.
\]
Accordingly, we distinguish between categorically grounded sequences $\boldsymbol{\sigma}$, and probabilistically grounded sequences $\boldsymbol{\tilde{\sigma}}$ using the tilde notation.
Finally, note that we use superscripts to indicate time steps in the sequence and subscripts to denote vector components. For instance, $\tilde{\sigma}^{(i)}_j$ denotes the $j$-th component of the probabilistic grounding of $\sigma$ at time step $i$.

\paragraph{Non-Markovian Reward Decision Processes}
In Reinforcement Learning (RL) \cite{sutton} the agent-environment interaction is generally modeled as a Markov Decision Process (MDP).
An MDP is a tuple $(S,A,t,r,\gamma)$, where $S$ is the set of environment \textit{states}, $A$ is the set of agent's \textit{actions}, $t: S \times A \times S \rightarrow [0,1]$ is the \textit{transition function}, $r:S \times A \rightarrow \mathbb{R}$ is the \textit{reward function}, and $\gamma \in [0,1]$ is the \textit{discount factor} expressing the preference for immediate over future reward.
In this classical setting, transitions and rewards are assumed to be Markovian – i.e., they are functions of the current state only.
Although this formulation is general enough to model most decision problems, it has been observed that many natural tasks are non-Markovian \cite{littman2017}.
A decision process can be non-markovian because markovianity does not hold on the reward function $r:(S\times A)^* \rightarrow \mathbb{R}$, or the transition function $t:(S \times A)^*\times S \rightarrow [0,1]$, or both. In this work we focus on Non-Markovian \textit{Reward} Decision Processes (NMRDP) \cite{restrainingBolts}.

\paragraph{Reward Machines}
Rather than developing new RL algorithms to tackle NMRDP, the research has focused mainly on how to construct Markovian state representations of NMRDP. An approach of this kind are the so called Reward Machines (RMs).
RMs are an automata-based representation of non-Markovian reward functions \cite{RM_journal}. Given a finite set of propositions $P$ representing abstract properties or events observable in the environment, a Reward Machine $RM$ is a tuple $(P, Q, R,  q_0, \delta, \lambda, L)$, where $P$ is the automaton alphabet, $Q$ is the set of automaton states, $R$ is a finite set of continuous reward values, $q_0$ is the initial state, $\delta: Q \times P \rightarrow Q$ is the automaton transition function, $\lambda: Q \rightarrow R$ is the reward function, and $L: S \rightarrow P$ is the labeling (or symbol grounding) function, which recognizes symbols in the environment states.
Let $\boldsymbol{s} = (s^{(1)}, s^{(2)}, ... , s^{(t)})$ be a sequence of states the agent has observed in the environment up to the current time instant $t$. This is transformed into a sequence of symbols $\boldsymbol{p} = (L(s^{(1)}), L(s^{(2)}), ... , L(s^{(t)}))$ by using the labeling function.
This string of symbols is processed by the Moore Machine $(P, Q, R, q_0, \delta, \lambda)$ so to produce an history-dependent reward (output) value at time $t$, $r^{(t)}$, and an automaton state at time $t$, $q^{(t)}$. 
The reward value can be used to guide the agent toward the satisfaction of the task expressed by the automaton, while the automaton state can be used to construct a Markovian state representation. In fact it was proven that the augmented state $(s^{(t)}, q^{(t)})$ is a Markovian state representation for the task expressed by the RM \cite{restr_bolts}.

\paragraph{Neural Reward Machines}
Neural Reward Machines (NRMs) are a probabilistic relaxation of standard Reward Machines, where the Moore machine is represented in matrix form, and input symbols, states, and rewards are \textit{probabilistically grounded}.  
Given a Moore machine \( (P, Q, R, q_0, \delta, \lambda) \) representing the task's reward structure—which is assumed to be known—we denote the transition and output (reward) functions in matrix form as \( \mathcal{T} \in \mathbb{R}^{|P| \times |Q| \times |Q|} \) and \( \mathcal{R} \in \mathbb{R}^{|Q| \times |R|} \), respectively.
NRMs assume that the labeling function \( L \) is unknown and must be approximated by a neural network \( sg \), which takes an environment state \( s \in S \) as input and outputs a probability distribution over symbols \( \tilde{p} \in \Delta(P) \), having trainable parameters \( \theta_{sg} \).  
The full model is formulated as follows:
\begin{equation} \label{eq:VRM_def}
\begin{array}{lll}
    \tilde{p}^{(t)} = sg(s^{(t)} ; \theta_{sg}) \qquad \qquad &
    \tilde{q}^{(t)} = \sum\limits_{j=1}^{j=|P|} \tilde{p}^{(t)}_j(\tilde{q}^{(t-1)} \cdot \mathcal{T}_{j} ) \qquad \qquad & 
    \tilde{r}^{(t)} = \tilde{q}^{(t)} \cdot \mathcal{R} \\ 
\end{array}
\end{equation}
The model is fully continuous and differentiable, allowing its parameters \( \theta_{sg} \) to be learned through gradient-based optimization on input-output target sequences.  
In particular, \cite{umili2024neuralrewardmachines} train the model on episodes \( (\boldsymbol{s}, \boldsymbol{r}) \) collected from interactions with the environment.

\section{Method}
\begin{figure}[t]
    \centering
    \includegraphics[width=0.9\linewidth]{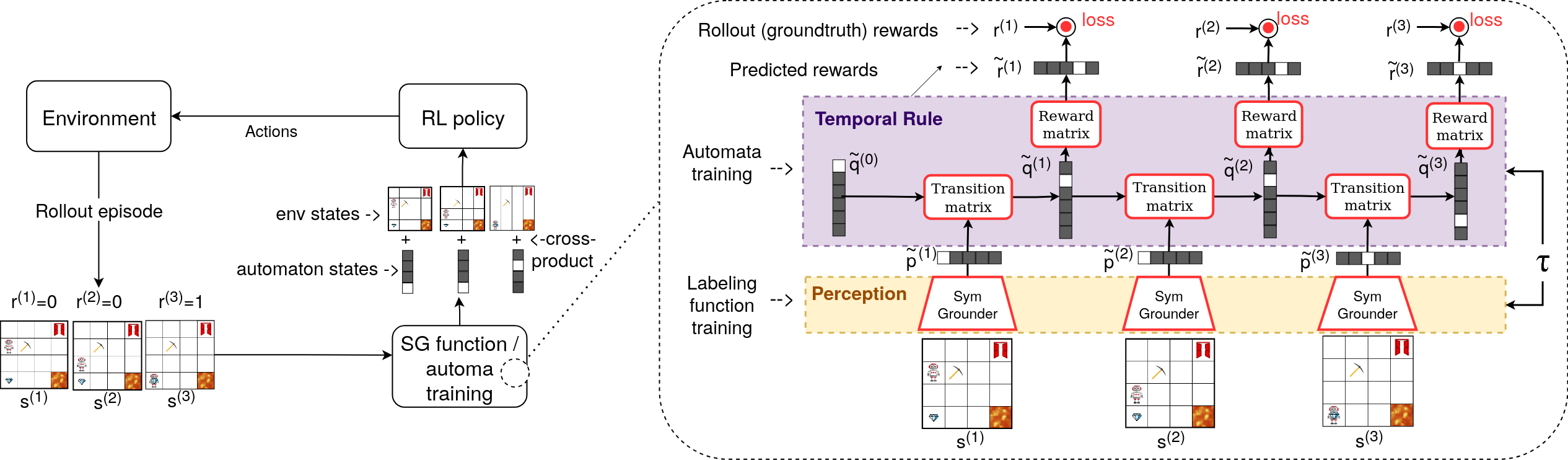}
    \caption{Integration of Fully Learnable Reward Machines within an RL schedule. The FLNRM is trained with episodes $(\boldsymbol{s}, \boldsymbol{r})$. At RL training time it produces the automaton states, that are used to augment the environment states so to mitigate non-markovianity. Models in red are trained through reward supervision.}
    \label{fig:FLNRM}
\end{figure}

\paragraph{Fully Learnable Reward Machines}
In this paper, we extend NRMs to be \textit{fully learnable}, and refer to our model as Fully Learnable Neural Reward Machines (FLNRM).
We assume that no prior knowledge is provided to the model and it must learn an approximation of both the labeling function and the Moore Machine from experience.
Since the task’s Moore machine specification is unknown, the number of required states and symbols is also unknown. We initialize the number of symbols to $|\hat{P}|$ and the number of states to $|\hat{Q}|$.
In contrast, the number of distinct reward values can be inferred through interaction with the environment, so we assume $|\hat{R}| = |R|$.
As a result, $|\hat{P}|$ and $|\hat{Q}|$ are the only two hyperparameters of our model.
The FLNRM model is shown in Figure  \ref{fig:FLNRM}, and it is formulated as follows
\begin{equation} \label{eq:transition_rnn}
\begin{array}{lll}
    \mathcal{T} = \text{softmax}(\theta_{\mathcal{T}} / \tau) & \mathcal{R} = \text{softmax}(\theta_{\mathcal{R}} / \tau) \\
    \tilde{p}^{(t)} = \text{softmax}( sg(s^{(t)} ; \theta_{sg})/ \tau) \qquad \qquad&
    \tilde{q}^{(t)}= \sum_{j=1}^{|\hat{P}|} \tilde{p}^{(t)}_j (\tilde{q}^{(t-1)} \cdot \mathcal{T}_j ) \qquad \qquad &
    \tilde{r}^{(t)} = \tilde{q}^{(t)} \cdot \mathcal{R}
\end{array}
\end{equation}
Our model has three learnable sets of parameters: $\theta_{sg}$, $\theta_{\mathcal{T}}$, and $\theta_{\mathcal{R}}$. Specifically, $\theta_{\mathcal{T}} \in \mathbb{R}^{|\hat{P}|\times |\hat{Q}| \times |\hat{Q}|}$ and $\theta_{\mathcal{R}} \in \mathbb{R}^{|\hat{Q}| \times |\hat{R}|}$ are matrices with the same dimensions as $\mathcal{T}$ and $\mathcal{R}$, respectively.
The matrices $\mathcal{T}$ and $\mathcal{R}$ are obtained by applying a softmax activation to the corresponding parameters. This activation ensures that $\mathcal{T}$ and $\mathcal{R}$ define valid probability distributions over the next state and output\footnote{Unless otherwise specified, the activation operates over the last dimension of each tensor. In this case, softmax ensures that each row of the matrix sums to one.}.
A temperature parameter $\tau$, with $0 < \tau \leq 1$, controls the sharpness of the softmax. When $\tau = 1$, the activation behaves normally; as $\tau$ approaches zero, the softmax approximates an argmax, and the model behaves increasingly like a deterministic finite state machine rather than a probabilistic one.
Deterministic behavior emerges when all rows in the transition and reward matrices become one-hot vectors.
We apply the same temperature-controlled activation to the symbol grounder network, so to smothly force the grounder to select only \textit{one} symbol with maximum probability at each time-step.
\paragraph{Integrating FLNRM with deepRL}
In this section, we describe how FLNRM is integrated with policy learning through RL in non-Markovian domains.  
As in standard RL, we consider an agent interacting with an unknown environment. At each time step \( t \), the agent takes an action \( a^{(t)} \), observes the current state \( s^{(t)} \), and receives a reward \( r^{(t)} \). The agent's objective is to learn a policy \( \pi: S \rightarrow A \) that maximizes the cumulative discounted reward:
\(
\sum_{t=0}^{\infty} \gamma^t r^{(t+1)}.
\)
We assume the reward signal is non-Markovian and can be modeled by a Reward Machine—namely, as the composition of a symbol perception function and a Moore machine.
As the agent explores the environment, we record each episode as a sequence of states \( \boldsymbol{s} \) and corresponding rewards \( \boldsymbol{r} \). At regular intervals, we use the collected experience to train the FLNRM parameters by minimizing the cross-entropy loss between the predicted reward sequence \( \boldsymbol{\tilde{r}} \) and the observed ground-truth rewards \( \boldsymbol{r} \).
Once the FLNRM has been trained, we use it to construct a history-dependent state representation to mitigate non-Markovianity. Specifically, we augment each environment state \( s^{(t)} \) with the probabilistically grounded machine state \( \tilde{q}^{(t)} \), and learn the policy over the augmented state space \(
\pi: S \times \Delta(\hat{Q}) \rightarrow A.\) A schema of this process is shown in Figure \ref{fig:FLNRM}.

\section{Experiments}

\begin{figure}[h!]
    \centering
    \label{fig:ImageRewards}
    
        \includegraphics[width=0.235\textwidth]{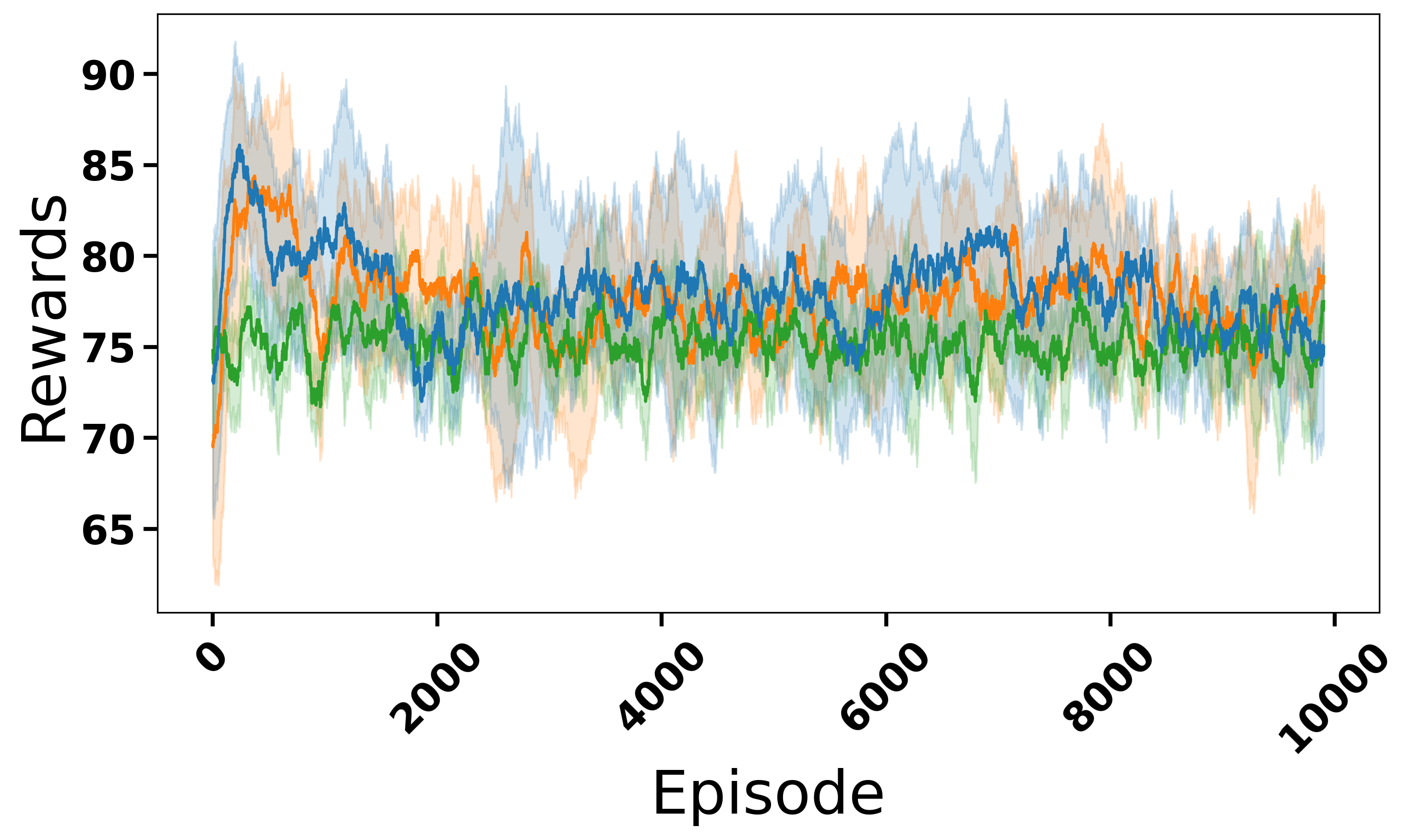}
    \hfill 
        \includegraphics[width=0.235\textwidth]{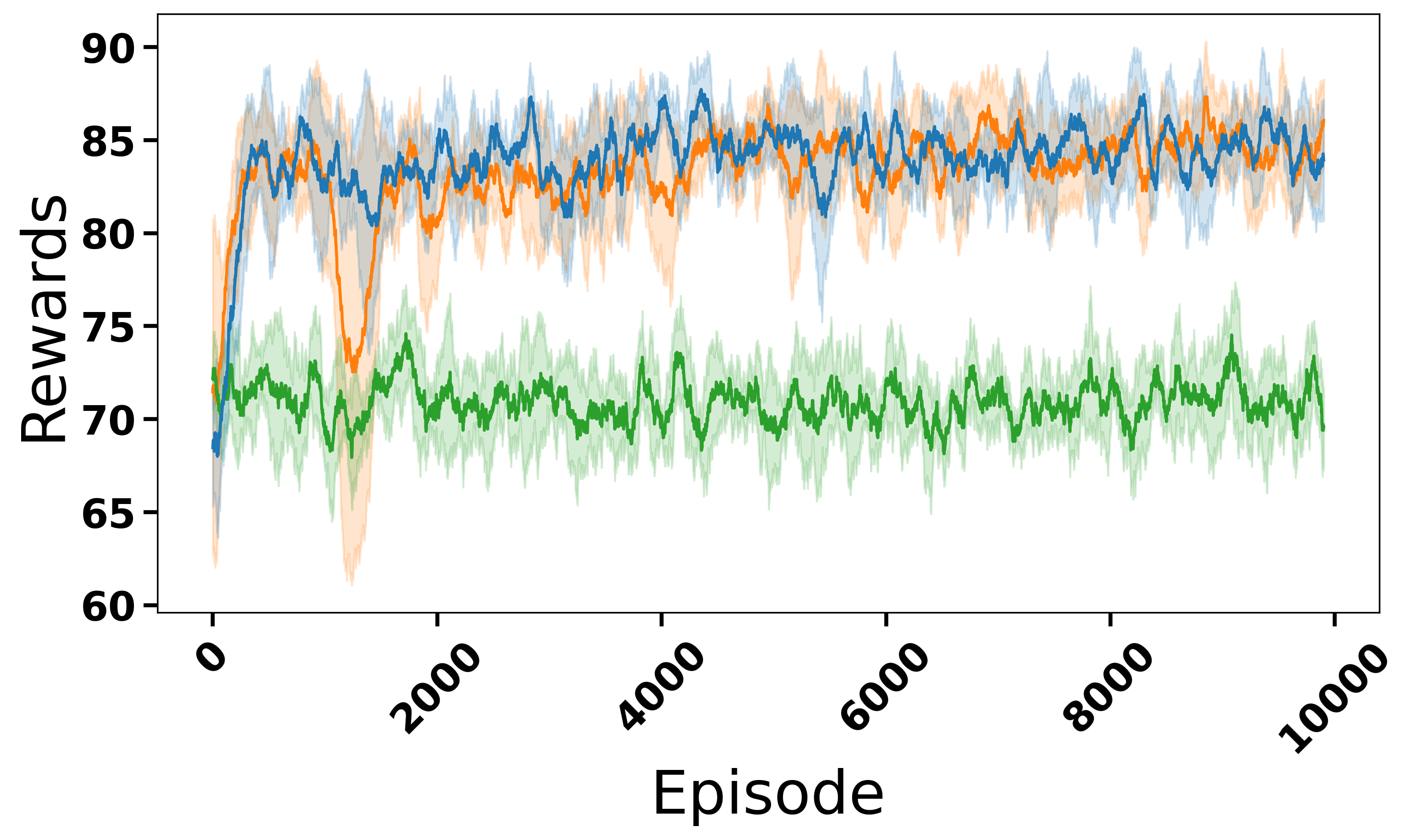}
    \hfill
        \includegraphics[width=0.235\textwidth]{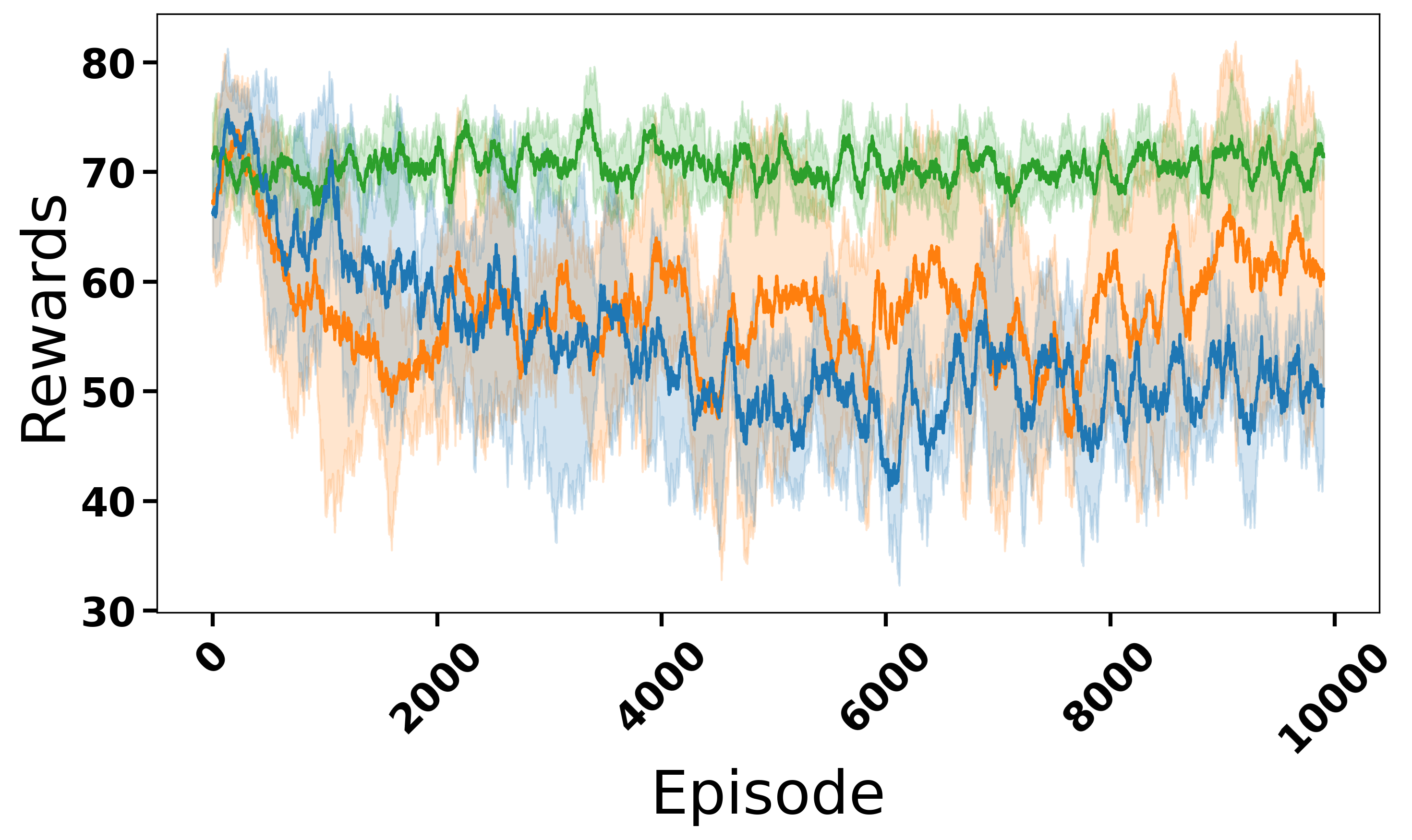}
    \hfill
        \includegraphics[width=0.235\textwidth]{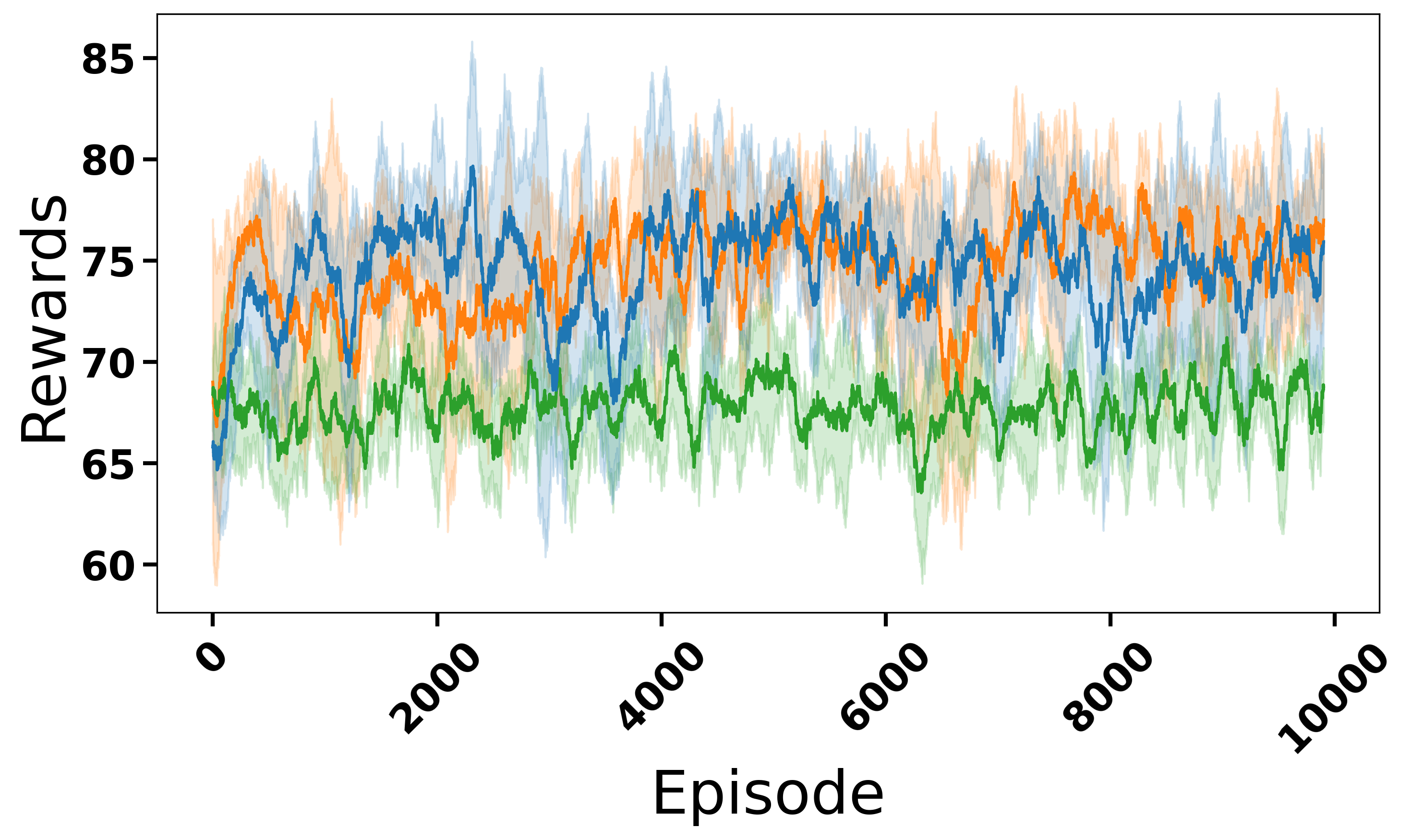}

    \vspace{-0.1em} 

        \includegraphics[width=0.235\textwidth]{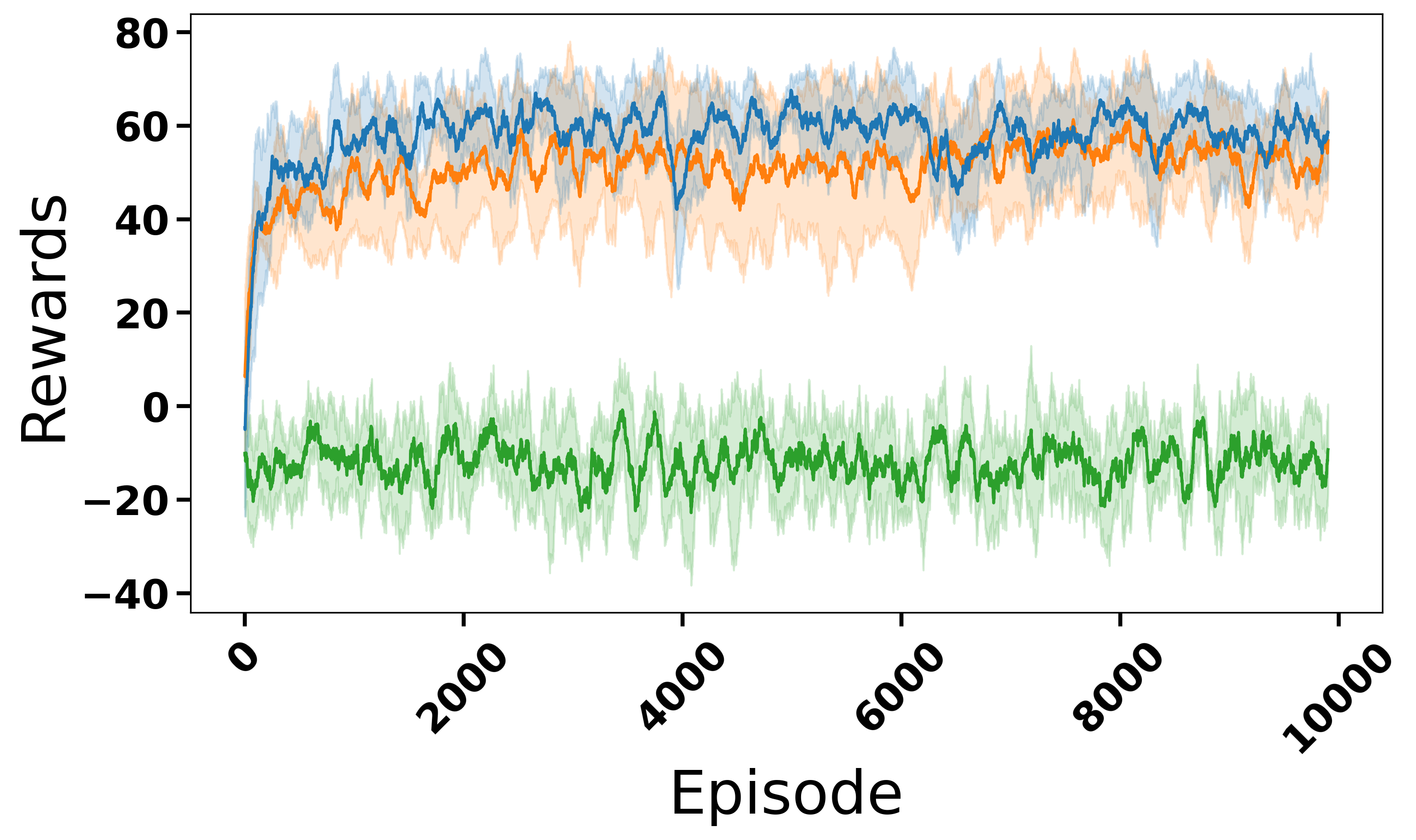}
    \hfill
        \includegraphics[width=0.235\textwidth]{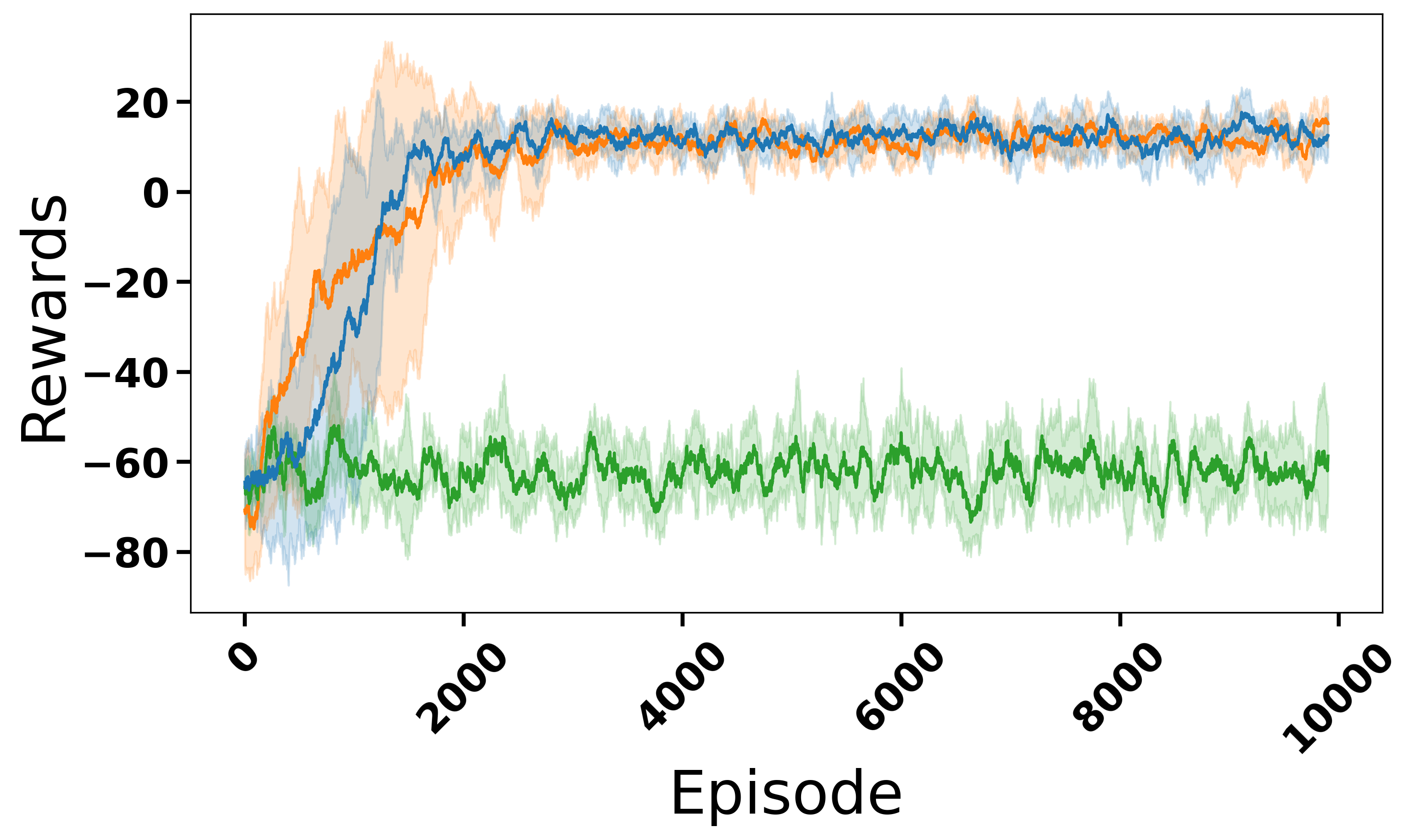}
    \hfill
        \includegraphics[width=0.235\textwidth]{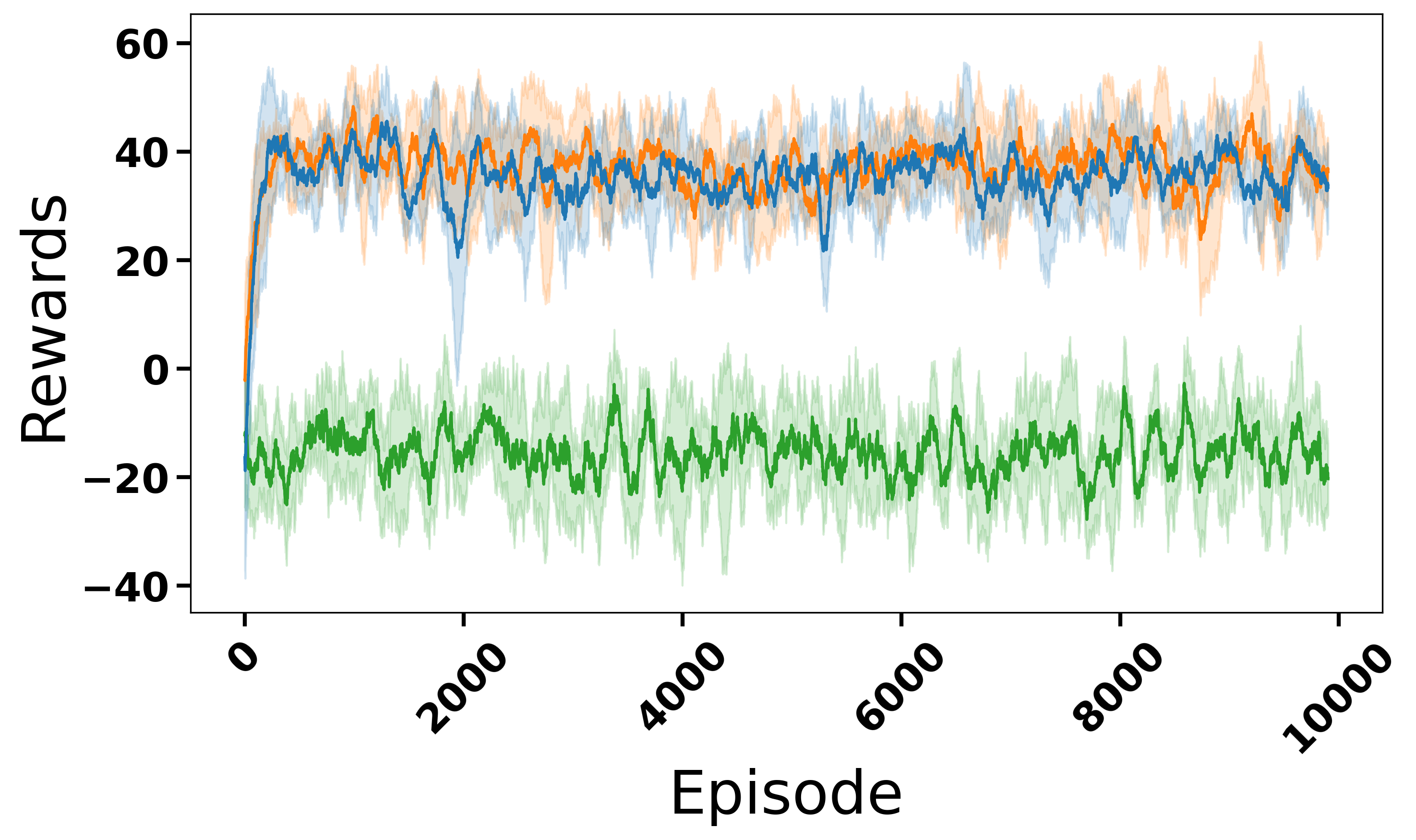}
    \hfill
        \includegraphics[width=0.235\textwidth]{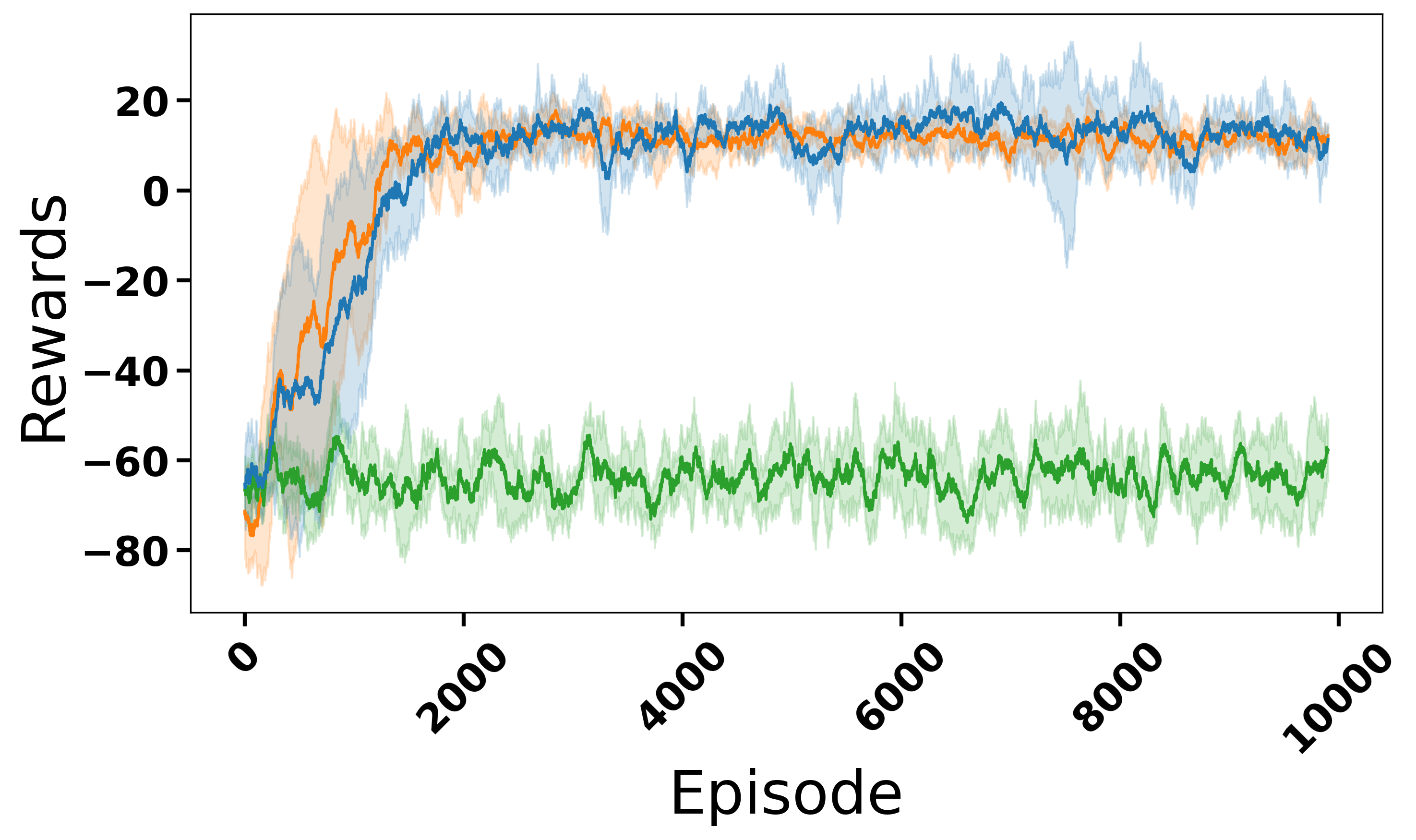}
\hrule
    \textbf{Legend}: \tikz[baseline={(0,-0.5ex)}]{\draw[NRM30, line width=2pt, line cap=round] (0,0) -- (0.3cm,0);} FLNRM with 30 states \tikz[baseline={(0,-0.5ex)}]{\draw[NRM5, line width=2pt, line cap=round] (0,0) -- (0.3cm,0);} FLNRM with 5 states \tikz[baseline={(0,-0.5ex)}]{\draw[RNN, line width=2pt, line cap=round] (0,0) -- (0.3cm,0);} RNN 
\hrule
        \includegraphics[width=0.235\textwidth]{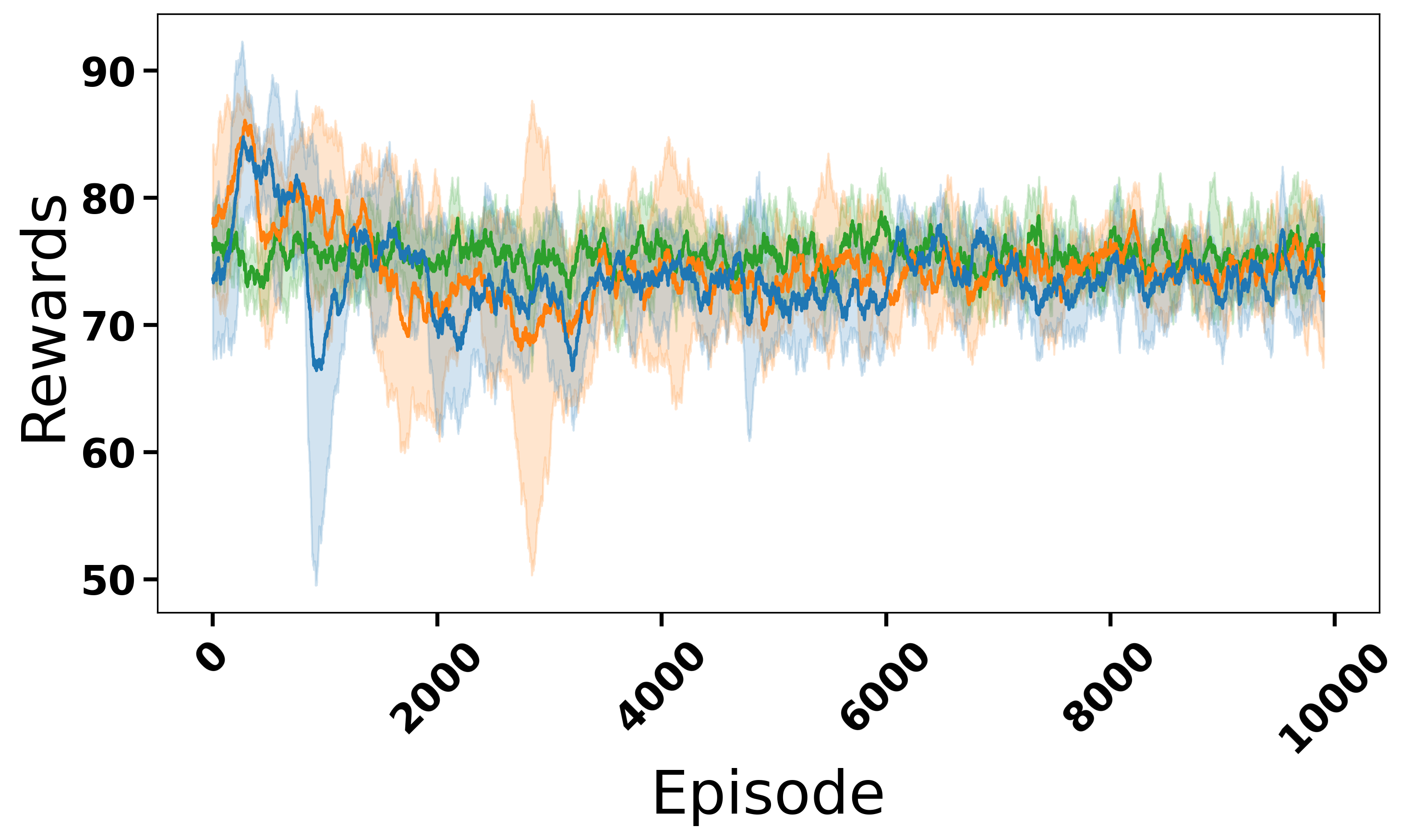}
    \hfill
        \includegraphics[width=0.235\textwidth]{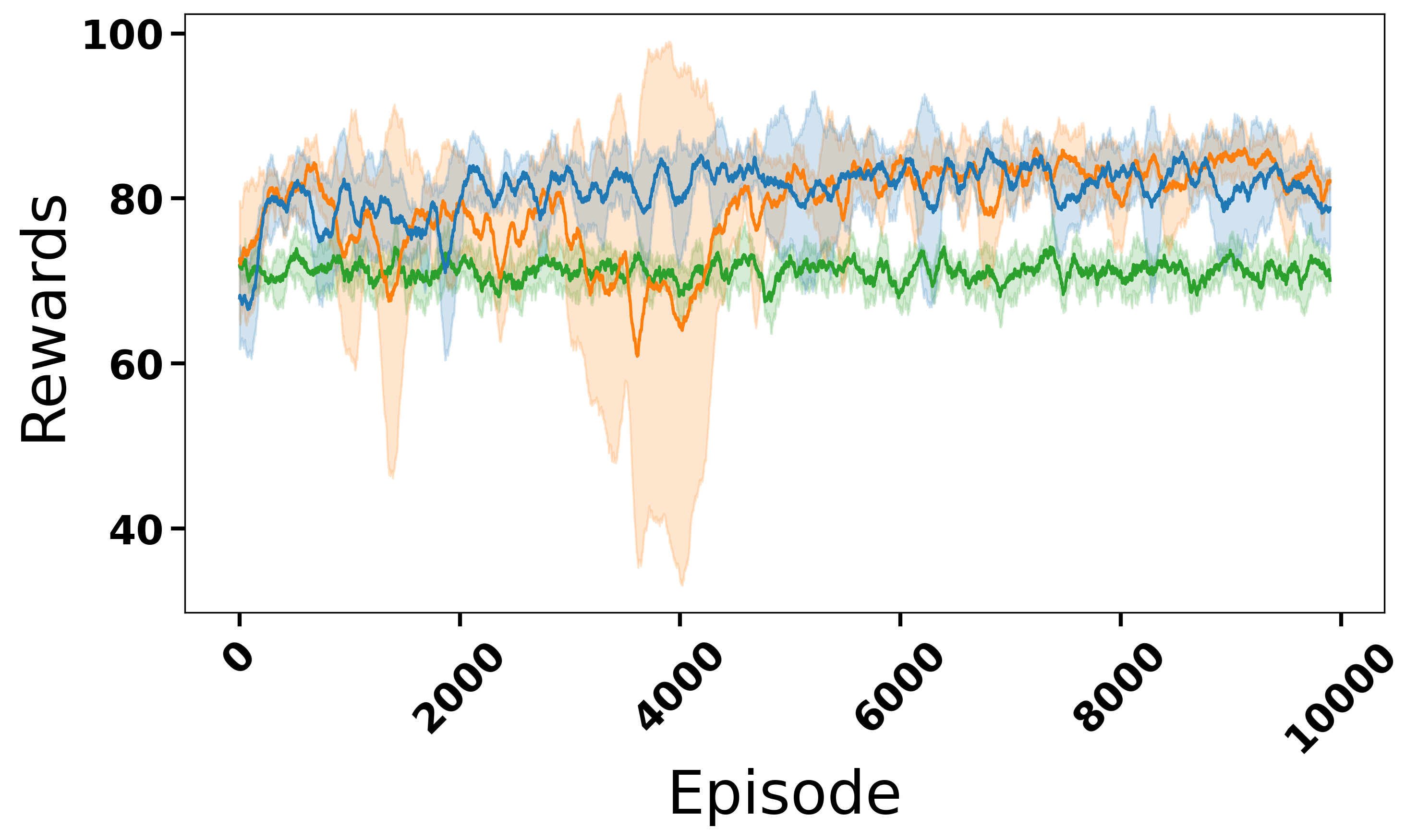}
    \hfill
        \includegraphics[width=0.235\textwidth]{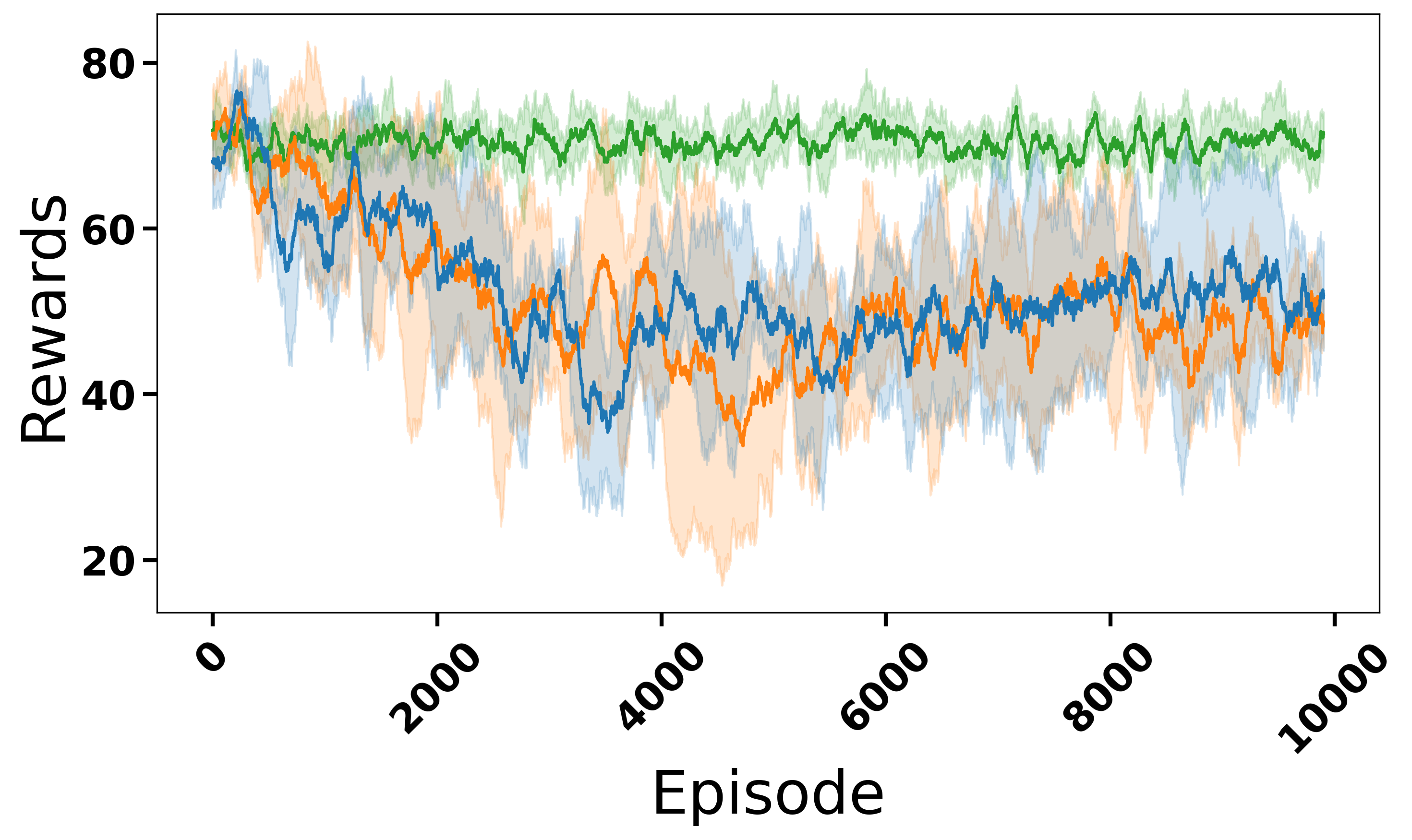}
    \hfill
        \includegraphics[width=0.235\textwidth]{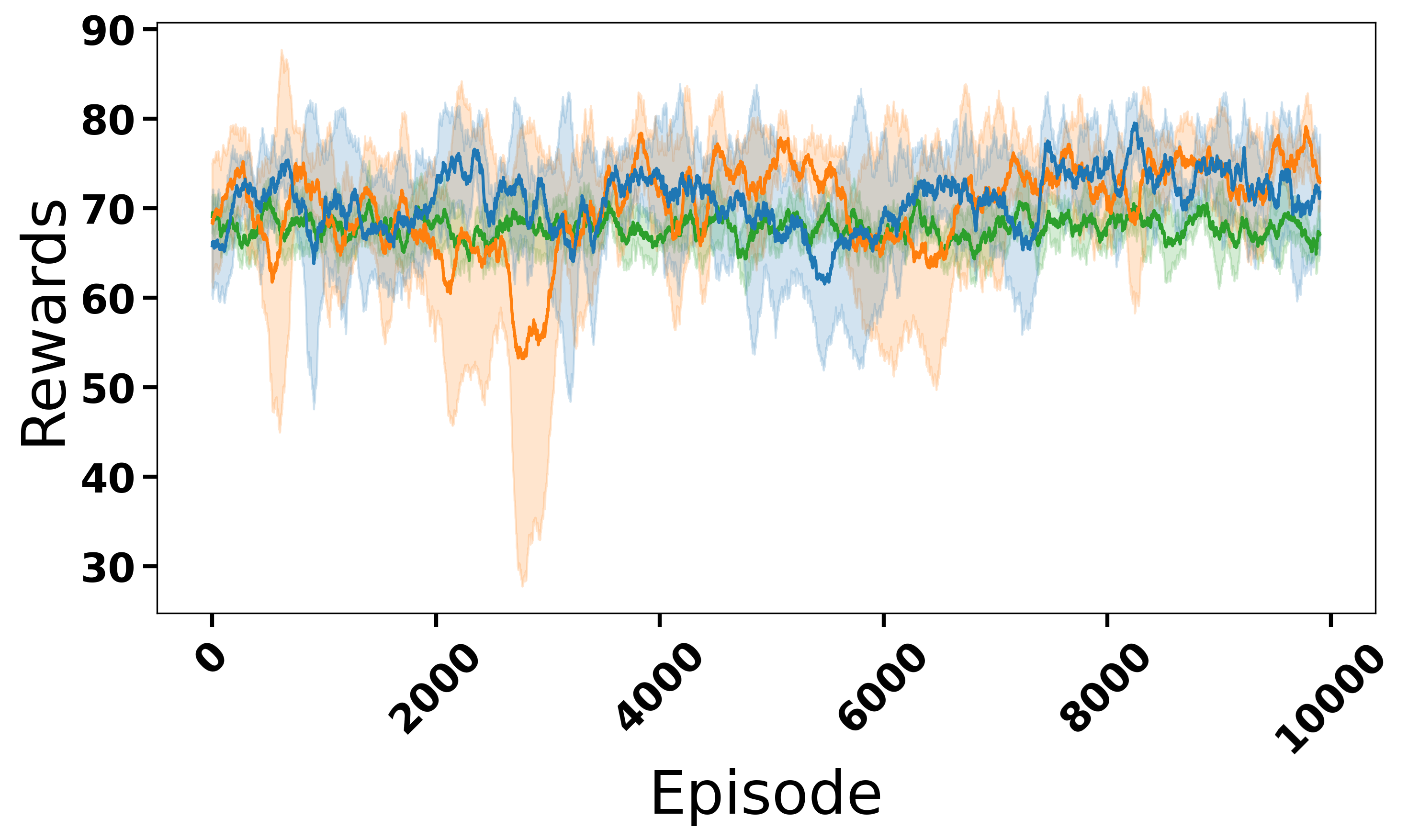}

    \vspace{-0.1em} 

        \includegraphics[width=0.235\textwidth]{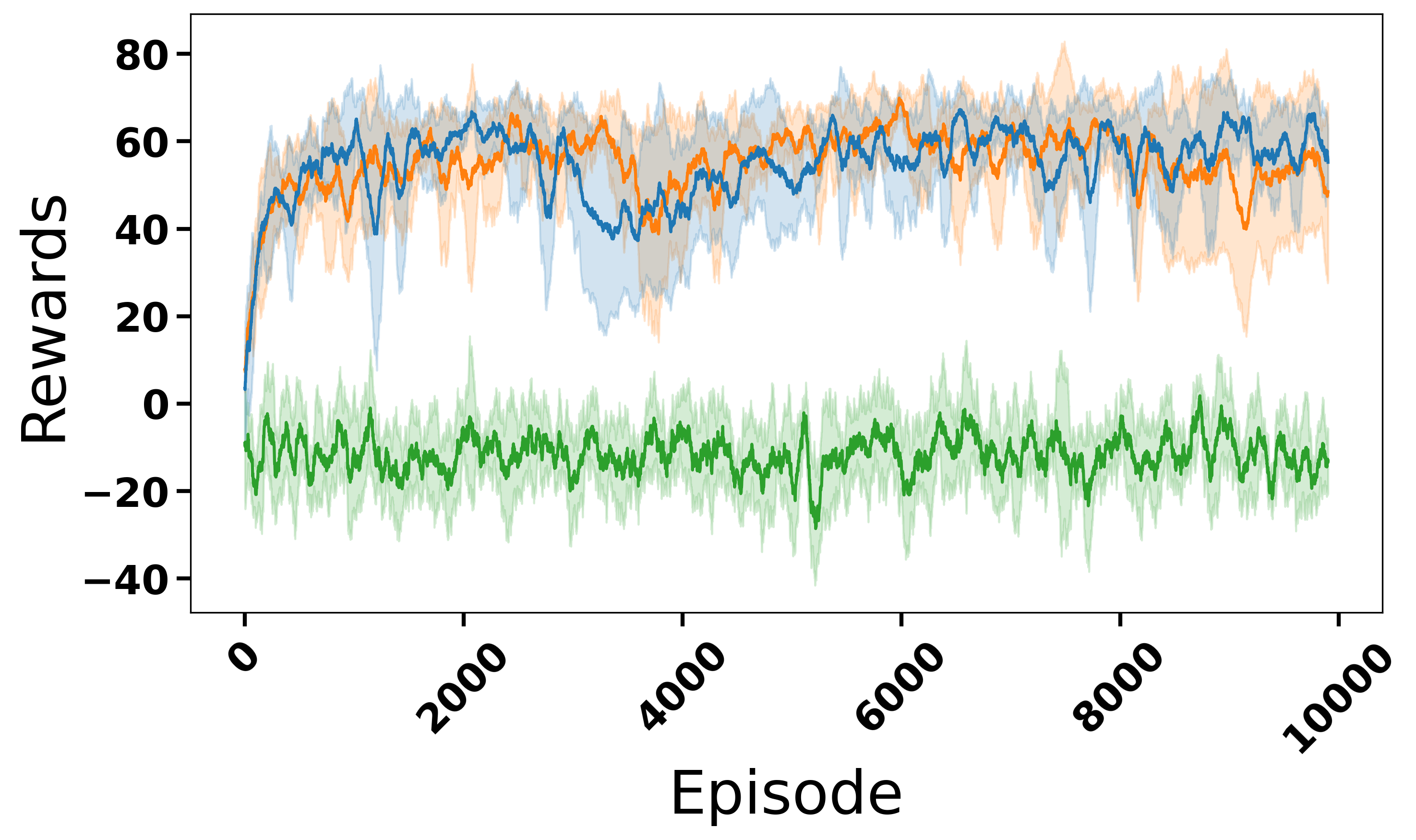}
    \hfill
        \includegraphics[width=0.235\textwidth]{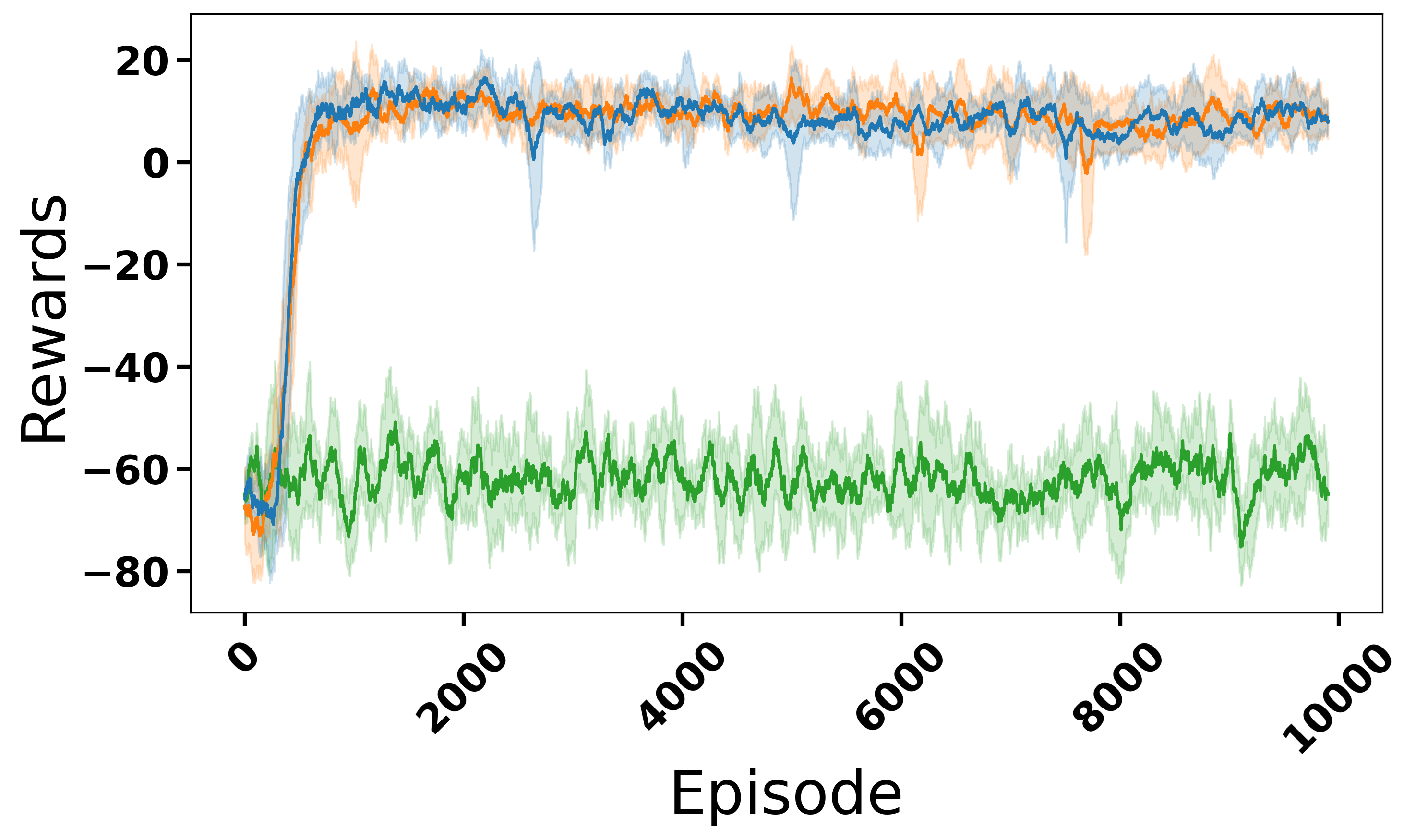}
    \hfill
        \includegraphics[width=0.235\textwidth]{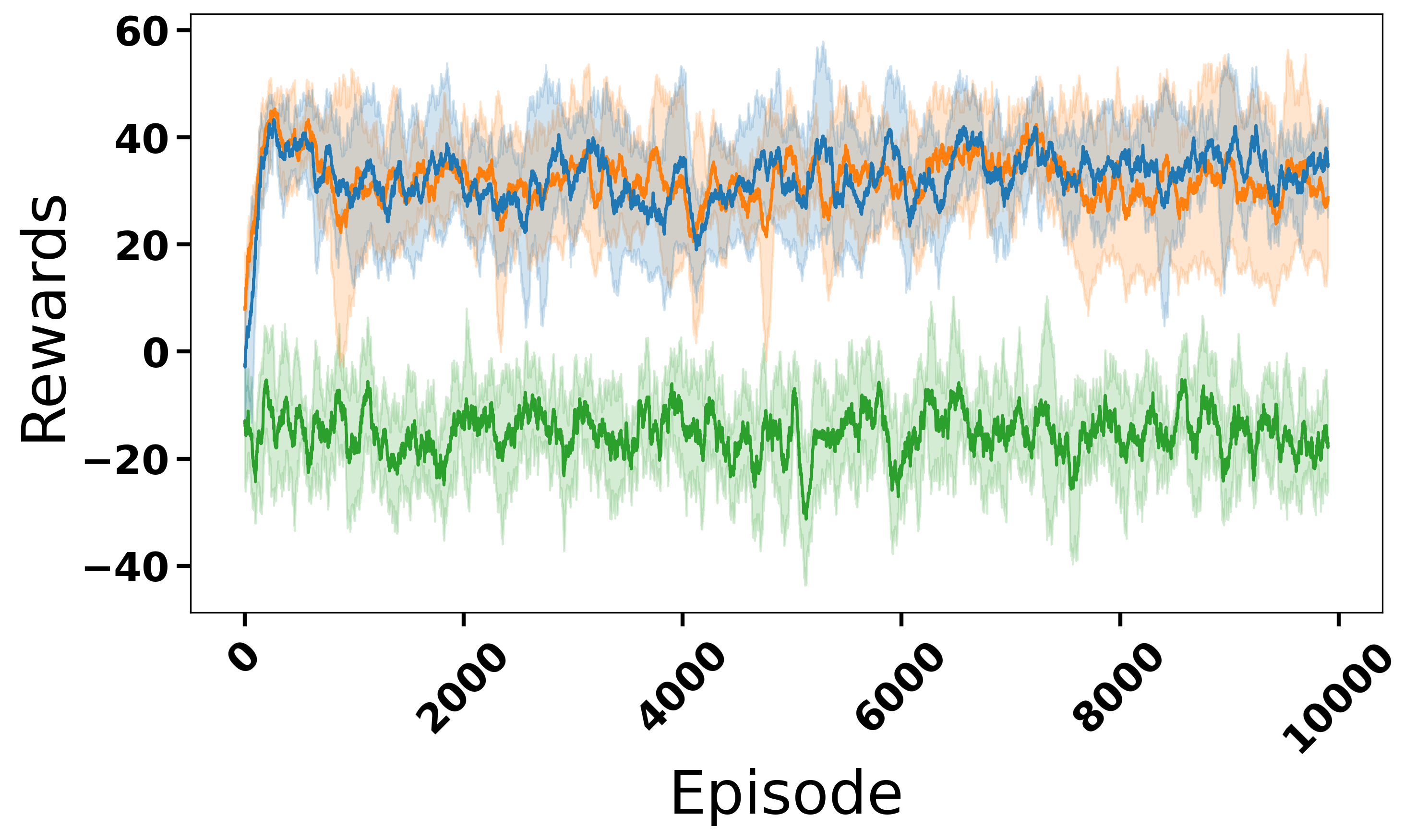}
    \hfill
        \includegraphics[width=0.235\textwidth]{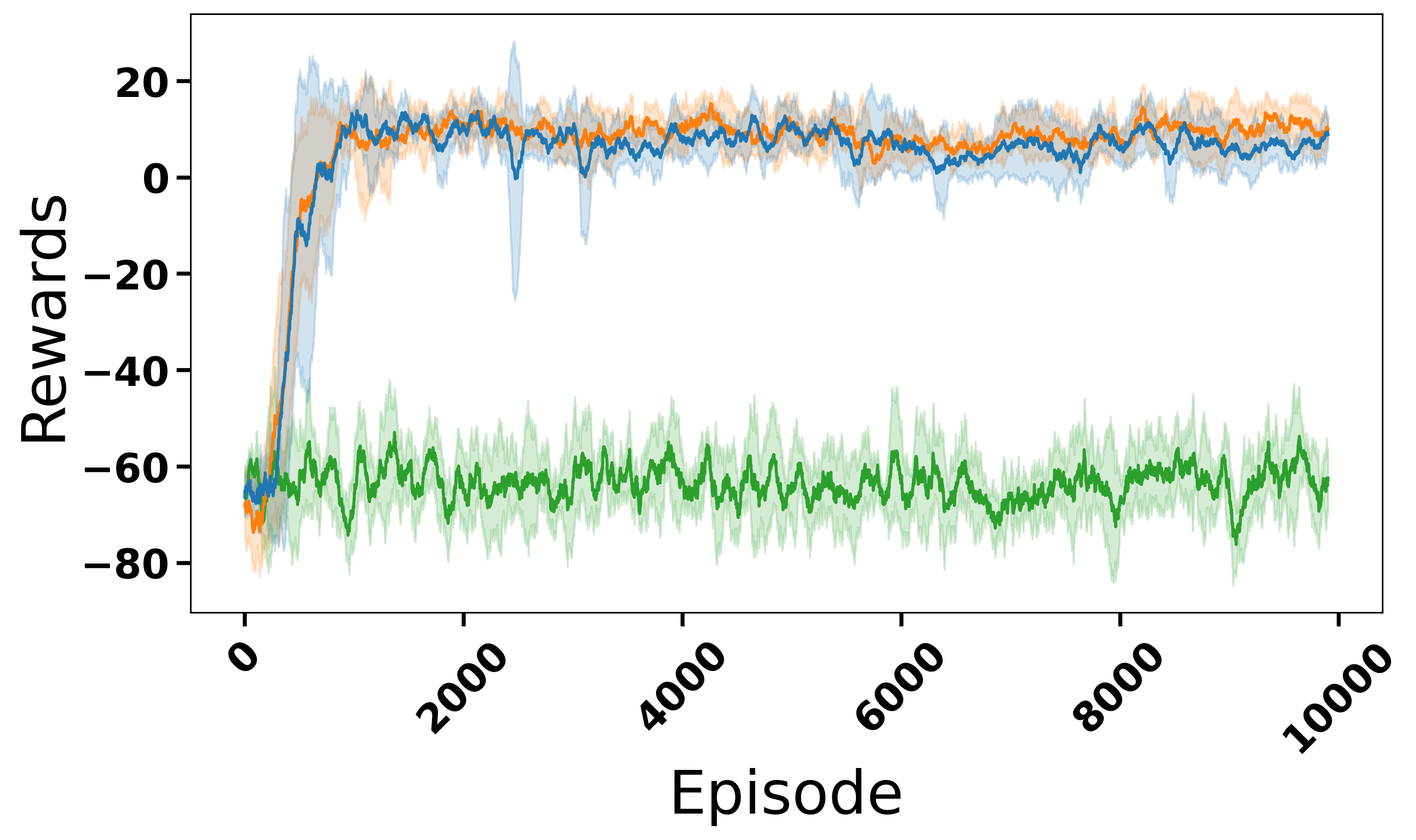}
    \caption{The first group of plots are about the image environment while the second bottom one is about the map environment. They show the training rewards obtained. First row: first class tasks. Second row: second class tasks.}
    \label{fig:MapRewards}
\end{figure}
We validate our framework by replicating the experimental setup presented in the NRM paper \cite{umili2024neuralrewardmachines}. Our implementation code is available at \href{https://github.com/KRLGroup/fully_learnable_neural_reward_machines}{Github} . In particular, we focus on navigation environments, where multiple items are present, and the agent must navigate among them so to satisfy a specific formula in Linear Temporal Logic over finite traces (LTLf) \cite{LTLf}. Two environments are designed to illustrate varying levels of difficulty in symbol grounding: 
(i) \textbf{Map Environment} – where the state is represented by a 2D vector indicating the agent’s current $(x, y)$ location;  
(ii) \textbf{Image Environment} – where the state consists of a $64 \times 64 \times 3$ pixel image depicting the agent within the grid.
For each of these two environments we tested two classes of temporal tasks, focusing on formula patterns commonly used in non-Markovian reinforcement learning~\cite{Icarte2022RewardMachines, Vaezipoor2021LTL2Action} and denoted as in~\cite{Menghi2021}:
(i) \textbf{first class} - includes tasks defined as conjunctions of \texttt{Visit} formulas (the agent must reach some items without a predefined order) and \texttt{Seq\_Visit} formulas (the agent must reach the items in a certain sequence).
(ii) \textbf{second class} - includes tasks defined as conjunctions of \texttt{Visit}, \texttt{Seq\_Visit}, and \texttt{Glob\_Avoid} formulas (the agent must always avoid certain items).
The complete list of formulas is reported in the Appendix.

\paragraph{Results}
We compare our method with RNN-based approaches using A2C \cite{MnihBMGLHSK16} as RL algorithm, $|\hat{P}|$ equal to the groundtruth number of symbols $|P|=5$, and $|\hat{Q}|$ equal to 5 and 30 states. Figures \ref{fig:MapRewards} show the training rewards obtained in both the image and map environments. For each task and method, we perform five runs with different random seeds. The results indicate that our method generally outperforms the baseline. Notably, the performance gap is most evident in the second class of tasks, which include the Global\_Avoidance constraint.
We attribute this to the strong and frequent feedback signals these clauses provide: violations trigger immediate and unambiguous negative rewards, which improve credit assignment and accelerate representation learning. 
All methods share the same hyperparameter settings for A2C, as well as for the neural networks used in the policy, value function, and feature extraction (the latter is only applied in the image environment), which are detailed in the appendix.
The results shows that the number of states will not affect much the quality of the model (the rewards are almost the same). Also changing the observation function only brings minor variations in the results. Indeed, for the same LTLf task, the reward trend is similar in both environments, despite one being based on images and the other on vector observations. This demonstrates that our method effectively handle different types of raw observations without any issues.

\section{Conclusions and Future Works}
In this paper, we extend NRMs into Fully Learnable NRMs, which learn an automaton representation of the RL task directly from raw observations and exploit it in real time to accelerate RL performance. Through extensive experimentation, we show that our method generally surpasses the performance of Deep RL baselines based on RNNs. Our method thus retains the same broad applicability and improved performance compared to DRL approaches, while being grounded in symbolic, explainable, and logic-based methods—combining the best of both worlds. One current limitation of our experiments is the assumption of knowing the ground-truth number of symbols—an unrealistic constraint in many real-world scenarios. In future work, we aim to test the framework with imprecise estimates of the number of symbols, further widening its applicability.
\begin{acknowledgments}
The work of Hazem Dewidar was carried out when he was enrolled in the Italian National Doctorate on Artificial Intelligence run by Sapienza University of Rome.
\noindent
This work has been partially supported by PNRR MUR project PE0000013-FAIR. 
\end{acknowledgments}

\section*{Declaration on Generative AI}
 During the preparation of this work, the author(s) used chat-GPT in order to: Grammar and spelling check. After using these tool(s)/service(s), the author(s) reviewed and edited the content as needed and take(s) full responsibility for the publication’s content. 

\bibliography{sample-ceur}

\begin{thebibliography}{25}
\expandafter\ifx\csname natexlab\endcsname\relax\def\natexlab#1{#1}\fi
\providecommand{\url}[1]{\texttt{#1}}
\providecommand{\href}[2]{#2}
\providecommand{\path}[1]{#1}
\providecommand{\DOIprefix}{doi:}
\providecommand{\ArXivprefix}{arXiv:}
\providecommand{\URLprefix}{URL: }
\providecommand{\Pubmedprefix}{pmid:}
\providecommand{\doi}[1]{\href{http://dx.doi.org/#1}{\path{#1}}}
\providecommand{\Pubmed}[1]{\href{pmid:#1}{\path{#1}}}
\providecommand{\bibinfo}[2]{#2}
\ifx\xfnm\relax \def\xfnm[#1]{\unskip,\space#1}\fi
\bibitem[{Ha and Schmidhuber(2018)}]{ha2018worldmodels}
\bibinfo{author}{D.~Ha}, \bibinfo{author}{J.~Schmidhuber},
\newblock \bibinfo{title}{Recurrent world models facilitate policy evolution},
\newblock in: \bibinfo{editor}{S.~Bengio}, \bibinfo{editor}{H.~Wallach}, \bibinfo{editor}{H.~Larochelle}, \bibinfo{editor}{K.~Grauman}, \bibinfo{editor}{N.~Cesa-Bianchi}, \bibinfo{editor}{R.~Garnett} (Eds.), \bibinfo{booktitle}{Advances in Neural Information Processing Systems}, volume~\bibinfo{volume}{31}, \bibinfo{publisher}{Curran Associates, Inc.}, \bibinfo{year}{2018}.
\bibitem[{Kapturowski et~al.(2019)Kapturowski, Ostrovski, Dabney, Quan, and Munos}]{kapturowski2019recurrent}
\bibinfo{author}{S.~Kapturowski}, \bibinfo{author}{G.~Ostrovski}, \bibinfo{author}{W.~Dabney}, \bibinfo{author}{J.~Quan}, \bibinfo{author}{R.~Munos},
\newblock \bibinfo{title}{Recurrent experience replay in distributed reinforcement learning},
\newblock in: \bibinfo{booktitle}{Proceedings of the 7th International Conference on Learning Representations (ICLR)}, \bibinfo{year}{2019}. \URLprefix \url{https://openreview.net/forum?id=r1lyTjAqYX}.
\bibitem[{Icarte et~al.(2022)Icarte, Klassen, Valenzano, and McIlraith}]{icarte2022reward}
\bibinfo{author}{R.~T. Icarte}, \bibinfo{author}{T.~Q. Klassen}, \bibinfo{author}{R.~A. Valenzano}, \bibinfo{author}{S.~A. McIlraith},
\newblock \bibinfo{title}{Reward machines: Exploiting reward function structure in reinforcement learning},
\newblock \bibinfo{journal}{Journal of Artificial Intelligence Research} \bibinfo{volume}{73} (\bibinfo{year}{2022}) \bibinfo{pages}{173--208}. \DOIprefix\doi{10.1613/JAIR.1.12440}.
\bibitem[{De~Giacomo and Vardi(2013)}]{degiacomo2013linear}
\bibinfo{author}{G.~De~Giacomo}, \bibinfo{author}{M.~Y. Vardi},
\newblock \bibinfo{title}{Linear temporal logic and linear dynamic logic on finite traces},
\newblock in: \bibinfo{booktitle}{Proceedings of the Twenty-Third International Joint Conference on Artificial Intelligence (IJCAI '13)}, \bibinfo{publisher}{AAAI Press}, \bibinfo{year}{2013}, pp. \bibinfo{pages}{854--860}.
\bibitem[{Umili et~al.(2024)Umili, Argenziano, and Capobianco}]{umili2024neuralrewardmachines}
\bibinfo{author}{E.~Umili}, \bibinfo{author}{F.~Argenziano}, \bibinfo{author}{R.~Capobianco}, \bibinfo{title}{Neural reward machines}, \bibinfo{year}{2024}. \URLprefix \url{https://arxiv.org/abs/2408.08677}. \href{http://arxiv.org/abs/2408.08677}{{\tt arXiv:2408.08677}}.
\bibitem[{Littman et~al.(2017)Littman, Topcu, Fu, Jr., Wen, and MacGlashan}]{littman2017}
\bibinfo{author}{M.~L. Littman}, \bibinfo{author}{U.~Topcu}, \bibinfo{author}{J.~Fu}, \bibinfo{author}{C.~L.~I. Jr.}, \bibinfo{author}{M.~Wen}, \bibinfo{author}{J.~MacGlashan},
\newblock \bibinfo{title}{Environment-independent task specifications via gltl},
\newblock \bibinfo{journal}{CoRR} \bibinfo{volume}{abs/1704.04341} (\bibinfo{year}{2017}). \URLprefix \url{http://arxiv.org/abs/1704.04341}.
\bibitem[{Camacho et~al.(2019)Camacho, Icarte, Klassen, Valenzano, and McIlraith}]{reward-machine-sheila}
\bibinfo{author}{A.~Camacho}, \bibinfo{author}{R.~T. Icarte}, \bibinfo{author}{T.~Q. Klassen}, \bibinfo{author}{R.~Valenzano}, \bibinfo{author}{S.~A. McIlraith},
\newblock \bibinfo{title}{Ltl and beyond: Formal languages for reward function specification in reinforcement learning},
\newblock in: \bibinfo{booktitle}{Proceedings of the Twenty-Eighth International Joint Conference on Artificial Intelligence (IJCAI-19)}, \bibinfo{publisher}{International Joint Conferences on Artificial Intelligence Organization}, \bibinfo{year}{2019}, pp. \bibinfo{pages}{6065--6073}.
\bibitem[{Giacomo et~al.(2021)Giacomo, Iocchi, Favorito, and Patrizi}]{restr_bolts}
\bibinfo{author}{G.~D. Giacomo}, \bibinfo{author}{L.~Iocchi}, \bibinfo{author}{M.~Favorito}, \bibinfo{author}{F.~Patrizi},
\newblock \bibinfo{title}{Foundations for restraining bolts: Reinforcement learning with ltlf/ldlf restraining specifications},
\newblock in: \bibinfo{booktitle}{Proceedings of the International Conference on Automated Planning and Scheduling}, volume~\bibinfo{volume}{29}, \bibinfo{year}{2021}, pp. \bibinfo{pages}{128--136}. \URLprefix \url{https://ojs.aaai.org/index.php/ICAPS/article/view/3549}.
\bibitem[{Gaon and Brafman(2020)}]{reward_machine_learning_1}
\bibinfo{author}{M.~Gaon}, \bibinfo{author}{R.~Brafman},
\newblock \bibinfo{title}{Reinforcement learning with non-markovian rewards},
\newblock \bibinfo{journal}{Proceedings of the AAAI Conference on Artificial Intelligence} \bibinfo{volume}{34} (\bibinfo{year}{2020}) \bibinfo{pages}{3980--3987}. \URLprefix \url{https://ojs.aaai.org/index.php/AAAI/article/view/5814}. \DOIprefix\doi{10.1609/aaai.v34i04.5814}.
\bibitem[{Xu et~al.(2021)Xu, Wu, Ojha, Neider, and Topcu}]{reward_machine_learning_2}
\bibinfo{author}{Z.~Xu}, \bibinfo{author}{B.~Wu}, \bibinfo{author}{A.~Ojha}, \bibinfo{author}{D.~Neider}, \bibinfo{author}{U.~Topcu},
\newblock \bibinfo{title}{Active finite reward automaton inference and reinforcement learning using queries and counterexamples},
\newblock in: \bibinfo{booktitle}{Machine Learning and Knowledge Extraction (CD-MAKE) 2021}, \bibinfo{year}{2021}, pp. \bibinfo{pages}{115--135}.
\bibitem[{Ronca et~al.(2022)Ronca, Licks, and Giacomo}]{reward_machine_learning_3}
\bibinfo{author}{A.~Ronca}, \bibinfo{author}{G.~P. Licks}, \bibinfo{author}{G.~D. Giacomo},
\newblock \bibinfo{title}{Markov abstractions for pac reinforcement learning in non-markov decision processes},
\newblock in: \bibinfo{booktitle}{Proceedings of the Thirty-First International Joint Conference on Artificial Intelligence (IJCAI 2022)}, \bibinfo{address}{Vienna, Austria}, \bibinfo{year}{2022}, pp. \bibinfo{pages}{3408--3415}. \URLprefix \url{https://doi.org/10.24963/ijcai.2022/473}. \DOIprefix\doi{10.24963/ijcai.2022/473}.
\bibitem[{Furelos-Blanco et~al.(2021)Furelos-Blanco, Law, Jonsson, Broda, and Russo}]{subgoalAutomaton}
\bibinfo{author}{D.~Furelos-Blanco}, \bibinfo{author}{M.~Law}, \bibinfo{author}{A.~Jonsson}, \bibinfo{author}{K.~Broda}, \bibinfo{author}{A.~Russo},
\newblock \bibinfo{title}{Induction and exploitation of subgoal automata for reinforcement learning},
\newblock \bibinfo{journal}{Journal of Artificial Intelligence Research} \bibinfo{volume}{70} (\bibinfo{year}{2021}) \bibinfo{pages}{1031--1116}. \URLprefix \url{https://doi.org/10.1613/jair.1.12372}. \DOIprefix\doi{10.1613/jair.1.12372}.
\bibitem[{Cai et~al.(2020)Cai, Xiao, Li, Li, and Kan}]{noisy_symbols_2020}
\bibinfo{author}{M.~Cai}, \bibinfo{author}{S.~Xiao}, \bibinfo{author}{B.~Li}, \bibinfo{author}{Z.~Li}, \bibinfo{author}{Z.~Kan},
\newblock \bibinfo{title}{Reinforcement learning based temporal logic control with maximum probabilistic satisfaction},
\newblock in: \bibinfo{booktitle}{2021 IEEE International Conference on Robotics and Automation (ICRA)}, \bibinfo{year}{2020}, pp. \bibinfo{pages}{806--812}.
\bibitem[{Verginis et~al.(2022)Verginis, Koprulu, Chinchali, and Topcu}]{noisy_symbols_2022}
\bibinfo{author}{C.~K. Verginis}, \bibinfo{author}{C.~Koprulu}, \bibinfo{author}{S.~Chinchali}, \bibinfo{author}{U.~Topcu},
\newblock \bibinfo{title}{Joint learning of reward machines and policies in environments with partially known semantics},
\newblock \bibinfo{journal}{CoRR} \bibinfo{volume}{abs/2204.11833} (\bibinfo{year}{2022}). \URLprefix \url{https://doi.org/10.48550/arXiv.2204.11833}. \DOIprefix\doi{10.48550/arXiv.2204.11833}.
\bibitem[{Li et~al.(2022)Li, Chen, Vaezipoor, Klassen, Icarte, and McIlraith}]{noisy_symbols_2022_shila}
\bibinfo{author}{A.~C. Li}, \bibinfo{author}{Z.~Chen}, \bibinfo{author}{P.~Vaezipoor}, \bibinfo{author}{T.~Q. Klassen}, \bibinfo{author}{R.~T. Icarte}, \bibinfo{author}{S.~A. McIlraith},
\newblock \bibinfo{title}{Noisy symbolic abstractions for deep rl: A case study with reward machines},
\newblock \bibinfo{journal}{CoRR} \bibinfo{volume}{abs/2211.10902} (\bibinfo{year}{2022}). \URLprefix \url{https://doi.org/10.48550/arXiv.2211.10902}. \DOIprefix\doi{10.48550/arXiv.2211.10902}.
\bibitem[{Kuo et~al.(2020)Kuo, Katz, and Barbu}]{ltl_no_grounding}
\bibinfo{author}{Y.~Kuo}, \bibinfo{author}{B.~Katz}, \bibinfo{author}{A.~Barbu},
\newblock \bibinfo{title}{Encoding formulas as deep networks: Reinforcement learning for zero-shot execution of ltl formulas},
\newblock in: \bibinfo{booktitle}{IEEE/RSJ International Conference on Intelligent Robots and Systems (IROS) 2020}, \bibinfo{year}{2020}, pp. \bibinfo{pages}{5604--5610}. \URLprefix \url{https://doi.org/10.1109/IROS45743.2020.9341325}. \DOIprefix\doi{10.1109/IROS45743.2020.9341325}.
\bibitem[{Hyde and Jr(2024)}]{hyde2024detectinghiddentriggersmapping}
\bibinfo{author}{G.~Hyde}, \bibinfo{author}{E.~S. Jr}, \bibinfo{title}{Detecting hidden triggers: Mapping non-markov reward functions to markov}, \bibinfo{year}{2024}. \URLprefix \url{https://arxiv.org/abs/2401.11325}. \href{http://arxiv.org/abs/2401.11325}{{\tt arXiv:2401.11325}}.
\bibitem[{Sutton and Barto(2018)}]{sutton}
\bibinfo{author}{R.~S. Sutton}, \bibinfo{author}{A.~G. Barto}, \bibinfo{title}{Reinforcement Learning: An Introduction}, \bibinfo{edition}{2nd} ed., \bibinfo{publisher}{The MIT Press}, \bibinfo{year}{2018}. \URLprefix \url{http://incompleteideas.net/book/the-book-2nd.html}.
\bibitem[{Giacomo et~al.(2019)Giacomo, Iocchi, Favorito, and Patrizi}]{restrainingBolts}
\bibinfo{author}{G.~D. Giacomo}, \bibinfo{author}{L.~Iocchi}, \bibinfo{author}{M.~Favorito}, \bibinfo{author}{F.~Patrizi}, \bibinfo{title}{Foundations for restraining bolts: Reinforcement learning with ltlf/ldlf restraining specifications}, \bibinfo{year}{2019}.
\bibitem[{Icarte et~al.(2022)Icarte, Klassen, Valenzano, and McIlraith}]{RM_journal}
\bibinfo{author}{R.~T. Icarte}, \bibinfo{author}{T.~Q. Klassen}, \bibinfo{author}{R.~A. Valenzano}, \bibinfo{author}{S.~A. McIlraith},
\newblock \bibinfo{title}{Reward machines: Exploiting reward function structure in reinforcement learning},
\newblock \bibinfo{journal}{Journal of Artificial Intelligence Research} \bibinfo{volume}{73} (\bibinfo{year}{2022}) \bibinfo{pages}{173--208}. \DOIprefix\doi{10.1613/JAIR.1.12440}.
\bibitem[{Giacomo and Vardi(2013)}]{LTLf}
\bibinfo{author}{G.~D. Giacomo}, \bibinfo{author}{M.~Y. Vardi},
\newblock \bibinfo{title}{Linear temporal logic and linear dynamic logic on finite traces},
\newblock in: \bibinfo{booktitle}{Proceedings of the Twenty-Third International Joint Conference on Artificial Intelligence (IJCAI '13)}, \bibinfo{publisher}{AAAI Press}, \bibinfo{year}{2013}, pp. \bibinfo{pages}{854--860}.
\bibitem[{Icarte et~al.(2022)Icarte, Klassen, Valenzano, and McIlraith}]{Icarte2022RewardMachines}
\bibinfo{author}{R.~T. Icarte}, \bibinfo{author}{T.~Q. Klassen}, \bibinfo{author}{R.~A. Valenzano}, \bibinfo{author}{S.~A. McIlraith},
\newblock \bibinfo{title}{Reward machines: Exploiting reward function structure in reinforcement learning},
\newblock \bibinfo{journal}{Journal of Artificial Intelligence Research} \bibinfo{volume}{73} (\bibinfo{year}{2022}) \bibinfo{pages}{173--208}. \DOIprefix\doi{10.1613/JAIR.1.12440}.
\bibitem[{Vaezipoor et~al.(2021)Vaezipoor, Li, Icarte, and McIlraith}]{Vaezipoor2021LTL2Action}
\bibinfo{author}{P.~Vaezipoor}, \bibinfo{author}{A.~C. Li}, \bibinfo{author}{R.~T. Icarte}, \bibinfo{author}{S.~A. McIlraith},
\newblock \bibinfo{title}{Ltl2action: Generalizing {LTL} instructions for multi-task reinforcement learning},
\newblock in: \bibinfo{booktitle}{Proceedings of the 38th International Conference on Machine Learning (ICML)}, \bibinfo{publisher}{PMLR}, \bibinfo{address}{Virtual Event}, \bibinfo{year}{2021}, pp. \bibinfo{pages}{10497--10508}.
\bibitem[{Menghi et~al.(2021)Menghi, Tsigkanos, Pelliccione, Ghezzi, and Berger}]{Menghi2021}
\bibinfo{author}{C.~Menghi}, \bibinfo{author}{C.~Tsigkanos}, \bibinfo{author}{P.~Pelliccione}, \bibinfo{author}{C.~Ghezzi}, \bibinfo{author}{T.~Berger},
\newblock \bibinfo{title}{Specification patterns for robotic missions},
\newblock \bibinfo{journal}{IEEE Transactions on Software Engineering} \bibinfo{volume}{47} (\bibinfo{year}{2021}) \bibinfo{pages}{2208--2224}. \URLprefix \url{https://doi.org/10.1109/TSE.2019.2945329}. \DOIprefix\doi{10.1109/TSE.2019.2945329}.
\bibitem[{Mnih et~al.(2016)Mnih, Badia, Mirza, Graves, Lillicrap, Harley, Silver, and Kavukcuoglu}]{MnihBMGLHSK16}
\bibinfo{author}{V.~Mnih}, \bibinfo{author}{A.~P. Badia}, \bibinfo{author}{M.~Mirza}, \bibinfo{author}{A.~Graves}, \bibinfo{author}{T.~P. Lillicrap}, \bibinfo{author}{T.~Harley}, \bibinfo{author}{D.~Silver}, \bibinfo{author}{K.~Kavukcuoglu},
\newblock \bibinfo{title}{Asynchronous methods for deep reinforcement learning},
\newblock \bibinfo{journal}{CoRR} \bibinfo{volume}{abs/1602.01783} (\bibinfo{year}{2016}). \URLprefix \url{http://arxiv.org/abs/1602.01783}. \href{http://arxiv.org/abs/1602.01783}{{\tt arXiv:1602.01783}}.

\end{thebibliography}

\appendix
\section{Experimental details}
\subsection{Task Formulas}
We selected 8 formulas as RL tasks, 4 of class 1 and 4 of class 2, that are detailed in Table \ref{tab:groundabilityfull}.
\begin{table*}[th!]
    \centering
    \begin{tabular}{
        >{\centering\arraybackslash}p{1cm}  
        >{\centering\arraybackslash}p{1cm}  
        >{\centering\arraybackslash}p{4cm}  
        >{\centering\arraybackslash}p{9cm}  
    }
        \toprule
        Task & Class & Pattern & Formula\\
        \midrule
        1 & 1 & Visit($a, b$) & F(a) $\wedge$ F(b)\\ \\
        2 & 1 & Visit($a, b, c$) & F(a) $\wedge$ F(b) $\wedge$ F(c)\\ \\
        3 & 1 & Sequenced\_Visit($a, b$) & F(a $\wedge$ F(b))\\ \\
        4 & 1 & Sequenced\_Visit($a, b$) + Visit($c$) & F(a $\wedge$ F(b)) $\wedge$ F(c)\\ \\
        5 & 2 & Visit($a, b$) + Global\_Avoidance($c$) & F(a) $\wedge$ F(b) $\wedge$ G($\neg$c)\\ \\
        6 & 2 & Visit($a, b$) + Global\_Avoidance($c, d$) & F(a) $\wedge$ F(b) $\wedge$ G($\neg$c) $\wedge$ G($\neg$d) \\ \\
        7 & 2 & Sequenced\_Visit($a, b$) + Global\_Avoidance($c$) & F(a $\wedge$ F(b)) $\wedge$ G($\neg$c) \\ \\
        8 & 2 & Sequenced\_Visit($a, b$) + Global\_Avoidance($c, d$) & F(a $\wedge$ F(b)) $\wedge$ G ($\neg$c) $\wedge$ G($\neg$d) \\ \\
        \bottomrule
    \end{tabular}
    \caption{Formulas selected as RL tasks}
    \label{tab:groundabilityfull}
\end{table*}
\subsection{Hyperparameters Setting}
The proposed model is designed to learn a variable number of latent states, denoted by \( \hat{Q} \). In our experiments, we evaluated performance under two configurations: \( \hat{Q} = 5 \) and \( \hat{Q} = 30 \).
\noindent
The recurrent neural network (RNN) component was configured as an LSTM of two layers (\texttt{num\_layers} = 2) and an output dimensionality of \texttt{rnn\_outputs} = 5. The size of the hidden state in the RNN was set to \texttt{rnn\_hidden\_size} = 50.
\noindent
For the Advantage Actor-Critic (A2C) architecture, the hidden layer size was fixed at \texttt{hidden\_size} = 120. The learning rate of the optimizer is set to \texttt{lr=0.0004} while the temperature value used is 0.5.

\end{document}